\documentclass[conference]{IEEEtran}
\IEEEoverridecommandlockouts
\usepackage{url}
\usepackage{cite}
\UseRawInputEncoding
\usepackage{amsmath,amssymb,amsfonts}
\usepackage{algorithmic}
\usepackage{graphicx}
\usepackage{textcomp}
\usepackage{xcolor}
\usepackage{lipsum,framed}
\def\BibTeX{{\rm B\kern-.05em{\sc i\kern-.025em b}\kern-.08em
    T\kern-.1667em\lower.7ex\hbox{E}\kern-.125emX}}

\usepackage{todonotes}

\begin{document}

% \title{Option Batch-Shuffle Trick: Answering MCQs on Telecommunication Documents using Large Language Models*\\
% % \thanks{Identify applicable funding agency here. If none, delete this.}
% }

\title{QMOS: Enhancing LLMs for Telecommunication with Question Masked loss and Option Shuffling\\

}

\author{
\IEEEauthorblockN{Blessed Guda\textsuperscript{*}, Gabrial Zencha Ashungafac\textsuperscript{*}, Lawrence Francis\textsuperscript{*}, Carlee Joe-Wong\textsuperscript{**}}
\IEEEauthorblockA{\textit{College of Engineering, Carnegie Mellon University} \\
% Kigali, Rwanda (Blessed Guda, Gabrial Zencha A., Lawrence Francis) \\
% Pittsburgh, USA (Carlee Joe-Wong) \\
blessedg@andrew.cmu.edu, gzenchaa@andrew.cmu.edu, lfrancis@andrew.cmu.edu, cjoewong@andrew.cmu.edu}
\thanks{\textsuperscript{*}These authors contributed equally to this work.}
\thanks{\textsuperscript{**}Mentor of the Team.}
}

% \thanks{\textsuperscript{*}These authors contributed equally to this work.}

% \author{\IEEEauthorblockN{Blessed Guda}
% \IEEEauthorblockA{\textit{College of Engineering} \\
% \textit{Carnegie Mellon University}\\
% Kigali, Rwanda \\
% blessedg@andrew.cmu.edu}
% \and
% \IEEEauthorblockN{Carlee Joe-Wong}
% \IEEEauthorblockA{\textit{College of Engineering} \\
% \textit{Carnegie Mellon University}\\
% Pittsburgh, USA \\
% cjoewong@andrew.cmu.edu}
% \and
% \IEEEauthorblockN{Gabrial Zencha A.}
% \IEEEauthorblockA{\textit{College of Engineering} \\
% \textit{Carnegie Mellon University}\\
% Kigali, Rwanda \\
% gzenchaa@andrew.cmu.edu}
% \and
% \IEEEauthorblockN{Lawrence Francis}
% \IEEEauthorblockA{\textit{College of Engineering} \\
% \textit{Carnegie Mellon University}\\
% Kigali, Rwanda \\
% lfrancis@andrew.cmu.edu}

% }

\maketitle
\begin{abstract}
% \tdCarlee{can make the author list more compact (e.g., put College of Engineering and CMU on the same line, and take out campus locations) and take out MCQs from the title to save space}
Large Language models (LLMs) have brought about substantial advancements in the field of Question Answering (QA) systems. These models do remarkably well in addressing intricate inquiries in a variety of disciplines. However, because of domain-specific vocabulary, complex technological concepts, and the requirement for exact responses, applying LLMs to specialized sectors like telecommunications presents additional obstacles. GPT-3.5 has been used in recent work,  to obtain noteworthy accuracy for telecom-related questions in a Retrieval Augmented Generation (RAG) framework. Notwithstanding these developments, the practical use of models such as GPT-3.5 is restricted by their proprietary nature and high computing demands. This paper introduces  QMOS, an innovative approach which uses a Question-Masked loss and Option Shuffling trick to enhance the performance of LLMs in answering Multiple-Choice Questions in the telecommunications domain. Our focus was on using open-source, smaller language models (Phi-2 and Falcon-7B) within an enhanced RAG framework. Our multi-faceted approach involves several enhancements to the whole LLM-RAG pipeline of finetuning, retrieval, prompt engineering and inference.
% Also due to the selection bias of LLMs in answering MCQs, we implement an inference time batch shuffling trick that enhances the accuracies of the model by margins of about 8\%. 
Our approaches significantly outperform existing results, achieving accuracy improvements from baselines of 24.70\% to 49.30\% with Falcon-7B and from 42.07\% to 84.65\% with Phi-2. Our code is open\footnote{\url{https://github.com/ai4africagroup/telcom_llm}}


\end{abstract}

\section{Introduction}
The field of Question Answering (QA) systems has witnessed significant advancements with the advent of large language models (LLMs) \cite{brown2020language}. These models have demonstrated remarkable capabilities in understanding and responding to complex queries across various domains. However, their application to specialized fields, such as telecommunications, presents unique challenges due to domain-specific terminology, intricate technical concepts, and the need for precise, accurate responses \cite{maatouk2024teleqna}.
Recent research has shown promising results in applying LLMs to telecom-specific QA tasks. 
% A notable study, Telco-RAG achieved 86\% accuracy using GPT-3.5 within a Retrieval Augmented Generation (RAG) framework for telecom-related questions \cite{bornea2024telco}. 
A notable study, Telco-RAG achieved accuracies around 75\% using GPT-3.5 within a Retrieval Augmented Generation (RAG) framework for telecom-related questions \cite{bornea2024telco}.
While this demonstrates the potential of LLMs in domain-specific applications, the computational demands and resource requirements of such models often pose significant challenges for practical, widespread implementation \cite{patterson2021carbon}. Also, the use of GPT-3.5 deemphasizes open-source AI, as the training data details and architecture of GPT-3.5 are not open. 
This paper explores an innovative approach to addressing multiple-choice questions (MCQs) in the telecommunications domain using \textit{open, small language models within an enhanced RAG framework.} 
Answering telecommunications standards questions presents a particularly challenging environment for LLMs due to its frequent use of unique abbreviations and rapidly evolving technologies and regulatory considerations \cite{maatouk2024teleqna}. To further complicate the problem, LLMs have been shown to demonstrate selection bias when answering MCQs which significantly degrades performance~\cite{pezeshkpour2023large}. In addition, due to rapid changes and data sparsity, large language models will require frequent fine-tuning to keep them up to date with latest telecommunications standard.
Our research aims to achieve competitive performance of larger models while maintaining efficiency and reducing computational costs. We focus on the Phi-2 \cite{microsoft2023phi2} and Falcon-7B \cite{almazrouei2023falconseriesopenlanguage}, which have shown promise in achieving competitive performance with significantly fewer parameters than their larger counterparts. 
% The telecommunications industry presents a particularly challenging environment for QA systems due to its rapidly evolving technology, complex network architectures, and regulatory considerations \cite{maatouk2024teleqna}.
% \tdCarlee{maybe elaborate here on why these are challenges for LLMs specifically. E.g., does rapid technology evolution mean the LLM needs to generalize well to new concepts, or facilitate frequent fine tuning to them?}
Our work addresses these challenges through a multi-faceted approach called \textit{\textbf{QMOS}} that combines several novel techniques:
\begin{enumerate}
    \item We leverage multiple embedding models to diversify and enrich the documents retrieved by the RAG system, potentially capturing a broader range of relevant information.
    \item Due to the high number of abbreviations used in Telecommunication standards, we enhance the dictionary of abbreviations used in \cite{bornea2024telco} to boost the hit rate of successful abbreviations. %This increased the hit rate of successful abbreviations from \textbf{63.74\%} to \textbf{95.16\%}.
    \item We meticulously design our prompts to guide the model to reason over the documents when selecting the answer. 
    \item We employ LoRA (Low-Rank Adaptation) fine-tuning on the Phi-2 model with \textit{a question-masked loss function} to efficiently adapt it to the telecommunications domain.
    \item We implement an innovative optimization technique: inference and train time option batch-shuffling, which enhances the accuracy of the Phi-2 model by eliminating bias in option position of correct answers. 
    % \item Crucially, our approaches significantly outperform baseline Phi-2 and Falcon 7B  RAG results (\textbf{24.70\% to 49.30\%} ) and Phi-2 models (\textbf{42.07\% to 81.65\%}). 
\end{enumerate}

% This comprehensive methodology aims to maximize the performance of small language models in a domain-specific context, allowing our approach to compete with larger, more resource-intensive systems.\\
% Our research contributes to the field of QA systems in telecommunication domain in several ways:
% \begin{enumerate}
%     \item It demonstrates the viability of using small language models in an enhanced RAG framework for domain-specific QA tasks, offering a more resource-efficient alternative to larger models.
%     \item It delves into the impact of prompt engineering on adapting small language models to specialized domains, by carefully constructing prompts that incorporate domain-specific terminology and context to enhance the model’s ability to address intricate technical questions within the telecommunications field.
%     \item It explores the effectiveness of LoRA fine-tuning in adapting small language models to specialized domains.
%     \item It explores the effectiveness of using multiple embedding models to diversify knowledge retrieval in RAG systems.
%     \item It introduces and analyzes the impact of inference time option batch-shuffling and the combination of multiple embedding models on retrieval quality and overall performance.
%     \item It provides a comprehensive evaluation of the proposed approach in the context of telecommunications MCQs, a domain that presents unique challenges for QA systems.
    
% \end{enumerate}

By focusing on small language models and employing these advanced techniques, our work addresses the growing need for efficient, scalable NLP solutions in specialized domains. This research has significant implications for developing cost-effective QA systems that can be deployed in resource-constrained environments while maintaining high accuracy.

% \textcolor{black}{The remainder of this paper is organized as follows: Section II reviews relevant literature on small language models, RAG Systems, and their applications in specialized domains. Section III details our proposed chunking strategy and RAG framework with the use of multiple embedding models. Section IV introduces our QMOS fine-tuning approach with the use of a Question Masked loss and Option Shuffling Trick. Section V presents our evaluations. Finally, section VI concludes the paper comparing the performance with the use of larger models.} 

% and IV details our proposed methodology, including the enhanced RAG framework, multiple embedding model integration, model prompting, model finetuning, and inference time batch shuffling.

\textcolor{black}{The paper is organized as follows: Section II reviews the current state of LLMs and QA in telecom. Section III outlines our methodology, including the RAG architecture, model fine-tuning, and batch shuffling. Section IV evaluates its effectiveness, and Section V concludes with future research directions}

\section{Related Work}

Small language models have emerged as a promising alternative to large models for domain-specific applications. Models like Phi-2, Falcon-7B and TinyLlama-1.1B\cite{zhang2024tinyllamaopensourcesmalllanguage}  offer advantages such as resource efficiency, faster inference times, and easier fine-tuning for specific domains \cite{hu2024minicpmunveilingpotentialsmall}. In \cite{piovesan2024telecomlanguagemodelslarge} Piovesan et al. (2024) conducted a comprehensive evaluation of the small language model Phi-2 in the telecommunications domain, comparing it to larger models like GPT-3.5 and GPT-4. Their findings demonstrate that Phi-2, despite being significantly smaller, achieved an overall accuracy of 52.30\% on the TeleQnA dataset\cite{maatouk2024teleqna}, compared to 67.29\% for GPT-3.5 and 74.91\% for GPT-4. Notably, when enhanced with Retrieval-Augmented Generation (RAG), Phi-2's performance in the challenging \textcolor{black}{"Standards Specifications"} category improved from 44.27\% to 56.63\%, nearly matching GPT-3.5's 56.97\%. In  another study, Ahmed et al.\cite{ahmed2024linguistic} conducted a comprehensive evaluation of several small language models, including Falcon 7B, in the telecommunications domain. Their findings demonstrate that Falcon 7B, despite having 7 billion parameters, achieved an overall accuracy of only 15.70\% on the TeleQnA dataset, significantly lower than larger models like GPT-3.5 (67.29\%) and GPT-4 (74.91\%). While there is limited study on improving the small language model for telecom application, there has been an exhausive study in improving the  \textcolor{black}{performance of} large language models. In \cite{soudani2024finetuningvsretrieval} Soudani et al show that finetuning and RAG are both feasible methods in improving the performance of language models. 

\textcolor{black}{These studies show that the small language models significantly lag begin the larger models in the telecom domain. In this research, we show that its possible to match and even outperform the performance of LLMs with SLMs in the telecom domain.}

% \textcolor{red}{Finetuning of large language models with limited resources have been proposed in \cite{hu2021lora} \& \cite{dettmers2024qlora}. Furthermore, In \cite{yepes2024financialreportchunkingeffective} it has been shown that finding the right chunk size  considerably improves RAG on financial data. While the effect of chunk sizes in telecommunications are investigated in \cite{bornea2024telco}, these investigations only consider LLMs and not  small language models. Specialized embedding models have been developed for various domains in a bid to improve RAG systems. Many RAG system rely on the massive text embedding leaderboard \cite{muennighoff2023mtebmassivetextembedding} for selection of an embedding model. However, there is no embedding that  fits all use cases, in \cite{khanna2024tabularembeddingmodeltem} , Khanna and Subedi shows how tabular RAG applications can be improved using finetuned embedding models.  }
% \tdCarlee{I don't think you have space for this, but you might give a short overview of related work on the methods you use to finetune the small models, e.g., saying that they don't consider the telecom use case}

\section{Proposed Methodology}
% \tdCarlee{In general, I think this section would benefit from a bit more clarity on which parts are your contributions (choices or designs specific to the telecom MCQ setting) and which are borrowed from past work}
\textcolor{black}{We propose  enhancing small language models like Phi-2 and Falcon-7B for telecom MCQ QA using an improved RAG framework with custom chunking, prompt engineering, and LoRA fine-tuning. We also introduce an option batch-shuffling technique to reduce selection bias, achieving accuracy comparable to larger models while maintaining efficiency}
\subsection{Retrieval-Augmented Generation Architecture}
Retrieval-Augmented Generation is an interesting technique used to enhance the performance of large language models on tasks where the required knowledge was not present in the training data~\cite{lewispatric2020}. 
RAG is achieved by integrating external knowledge sources in the prompt from which the LLM’s result is generated. In question answering tasks, questions are augmented with contextually relevant texts from external documents when creating prompts.

\subsubsection{Splitting Documents into Chunks}
For RAG, we only need relevant parts of a large number of documents, and this  requires a search over these documents. 
% To facilitate this search, we create several smaller chunks of texts from these documents, ensuring that these chunks contain meaningful information about the documents from which they where extracted.
% One approach for this split is to obtain all the sentences in the documents such that each sentence becomes a chunk. Since several contiguous sentences from a document are required to convey certain information, it is not ideal to have chunks of individual sentences.
A common approach being employed by commercial tools for RAG such as LlamaIndex\cite{LlamaIndex} and LangChain\cite{LangChain} splits the document such that each chunk contains a certain number of characters with an overlap between chunks. The parameters 'chunk-size' and 'chunk-overlap' are often adjusted such that the split leads to meaningful chunks ('chunk-size' specifies the number of characters in a chunk of text while 'chunk-overlap' specifies the overlap of texts between two contiguous chunks). We employed \textit{a custom document splitting strategy }where we first split the documents into individual sections, excluding the table of contents. Each section is further split into chunks containing ‘chunk-size’ characters (in our setting, chunk-size = 1024). Additionally we ensured that each chunk begins with the heading of the document section to which it belongs by prepending this heading to each \textcolor{black}{chunk. This} chunking strategy is inspired by our analysis of the structure of the 3GPP standard documents. Also, we observe that the first few pages of the documents containing sections like the title page, scope, references  and table of contents are not very informative and as such we do not include those sections when creating the chunks. Having obtained the necessary chunks, we proceed to create vector embeddings of these chunks.
% \tdCarlee{Maybe point out that this is inspired by the specific structure of telecom documents} 

\subsubsection{Creating chunk embeddings}
To enable similarity search, we created associated embeddings for each chunk using an embedding model, as shown in \textcolor{black}{Figure~\ref{fig:chunk-retrieval}}. The choice of embedding model was influenced by models on the Massive Text Embedding Benchmark (MTEB) leaderboard ~\cite{muennighoff2023mtebmassivetextembedding}. We opted for the best performing and easily accessible models, favouring the use of the models \textcolor{black}{"stella\_en\_400M\_v5" and "gte-Qwen2-1.5B-instruct"}~\cite{li2023towards} with 400M and 1.5B parameters respectively. We used these text embedding models from the 'Sentence Transformer' library ~\cite{reimers-2019-sentence-bert} and sped up the  embedding process by batching the chunks, using a batch size of 64. The chunks and their corresponding embeddings were saved to disk to be used for context retrieval for the question answering task.
% \begin{figure}
%     \centering
%     \includegraphics[width=0.9\linewidth]{chunking_drawio_1.png}
%     \caption{Document splitting and chunk embedding process}
%     \label{fig:doc-splits}
% \end{figure}

\subsubsection{Chunk retrieval}
\begin{figure*}[!t]
    \centering
    \includegraphics[width=0.9\textwidth]{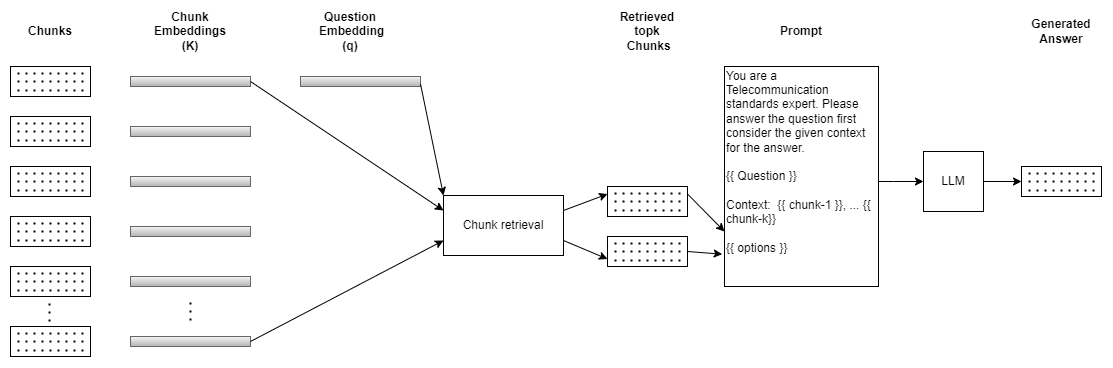}
    \caption{Chunk retrieval process: The top k chunks are retrieved using the KNN algorithm. Scores for each chunk is obtained from the dot product \textcolor{black}{$q\cdot K^T$}. The retrieved chunks are then used to create the input prompt to the LLM}
    \label{fig:chunk-retrieval}
\end{figure*}

For the chunk retrieval process, we employed a k-Nearest Neighbors (KNN) approach on the similarity scores between a question/query embedding and the chunk embeddings~\cite{9065747}. The use of a dot product similarity score implies that higher scores denote higher similarity. Hence, we retrieve the top $k$ similar chunks for a given question/query. The number of retrieved chunks is chosen such that the context length of the language model is not exceeded when creating the prompt. We used the top 2 chunks retrieved with each embedding model. Figure~\ref{fig:chunk-retrieval} shows how chunks are retrieved to form a context in the input prompt to an LLM. 

Additionally, we employed the use of the BM25 ~\cite{10.1561/1500000019} algorithm which is a statistical approach for information retrieval that measures similarity based on the frequency of terms from the query that appear in the chunks. The motivation behind this is to ensure that the retrieval also includes chunks containing specific terms used in the query but are not necessarily enforced in the neural embedding models.This enabled us to create a context that consists of chunks from 2 embedding models (stella\_en\_400M\_v5  and gte-Qwen2-1.5B-instruct~\cite{li2023towards}) and BM25~\cite{10.1561/1500000019}.

\subsubsection{Model Prompt}
Prompt engineering is crucial in RAG-based LLM systems as it significantly enhances their performance and reliability. By carefully crafting prompts, engineers can guide LLMs to produce more accurate, relevant, and contextually appropriate responses without the need for extensive finetuning ~\cite{marvin2023prompt}. This is particularly important in RAG systems, where the integration of external knowledge sources in the prompts helps mitigate issues like hallucinations and factual inaccuracies ~\cite{vatsal2024surveypromptengineeringmethods}. In designing our prompt we make the following considerations:
\begin{itemize}
    \item \textbf{Question Repitition: } we draw from the observation of ~\cite{bornea2024telco}, which showed that in answering Telecommunication questions, repeating the question before and after the contexts helps make the Phi-2 model reason over the contexts for the answer. 
    \item  \textbf{Enhanced Abbreviation Expansion: }  As we noticed that a lot of the questions in the TeleQnA dataset are about abbreviations, we also decide to include the abbreviations in the prompt like ~\cite{bornea2024telco}. However, we notice that the method in ~\cite{bornea2024telco} missed a lot of abbreviations because of the insufficient dictionary of abbreviations. This is because the abbreviation dictionary used was generated by only considering the Vocabulary for 3GPP specifications document\cite{vocab3GPP}. We expand this dictionary by searching for other abbreviations in the "\textit{Definitions of terms, symbols and abbreviations}"  sections of all the documents. With this enhanced abbreviation dictionary, we are able to achieve a hit rate of \textbf{95.16\%} in the test set which is a significant improvement over the \textbf{63.74\%} achieved using the approach in ~\cite{bornea2024telco}. 

    \item  \textbf{Model Prompt format}: We consider how the prompts in the training phase of the model were structured. This is crucial because presenting the model with prompts structured based on how it's trained can help the model perform better. 
    
\end{itemize}

% In designing our prompt for the Phi-2 model, we use the following 
With the above considerations, we design our prompt for Phi-2 as:

\begin{framed}
Instruct: **\textit{Question}**
 
 Abbreviations:

**\textit{abbreviation: full form}**

Considering the following retrieved contexts

context 1:  context...

context 2:  context...

**\textit{Question}**

** option 1: ....

** option 2: ....

Output : 
\end{framed}

\subsubsection{Falcon-7B}
We make slight modifications to the prompt for the Falcon 7B based on its poor performance on the TeleQnA dataset\cite{ahmed2024linguistic}.  We observed that when provided with the options, the model does not do well therefore, we do not provide the options to the model.

\begin{framed}
Youre a Telecommunication standards expert. Please answer the question  first consider the given context for the answer. 

 **\textit{Question}**
 
 Abbreviations:

**\textit{abbreviation: full form}**

Considering the following retrieved contexts

context 1:  context...

context 2:  context...

**\textit{Question}**

\end{framed}

From the answer sentence generated by the model, we compute the embeddings of the generated answer sentence and take the answer option that has the highest similarity. We use a simple sentence BERT model\cite{DBLP:journals/corr/abs-1908-10084} for this answer extraction. 
 
 % \begin{itemize}
% Instruct: **\textit{Question}**
%  Abbreviations:
% **\textit{abbreviation: full form}**
% \end{itemize}

% How is the prompt defined for the model?

% [Display the prompt used for a sample question]

\subsection{QMOS Finetuning approach}
\subsubsection{Phi-2 Finetuning}
To finetune the Phi-2 model, we used LoRA (Low-Rank Adaptation) ~\cite{hu2021lora}  fine-tuning approaches instead of full finetuning due to the size of the model and the small size of the training data. We explored both LoRa and Quantized LoRa (QLoRa) ~\cite{dettmers2024qlora} fine-tuning. We did not explore fine-tuning the Falcon 7B model, due to its large size which exceeds our computational budget.
% \tdCarlee{Did you also finetune the Falcon model? If not, why not?}

\subsubsection{QLoRa}
QLoRA (Quantized Low-Rank Adaptation) is a parameter-efficient fine-tuning technique for large language models. It combines quantization and low-rank adaptation to significantly reduce memory requirements and computational costs. In QLoRA, the base model's weights are quantized to 4-bit precision and frozen. Then, small trainable "adapter" layers are added using low-rank decomposition. We envision that these adapters would capture the Telecommunication MCQ reasoning ability during fine-tuning while keeping most of the model fixed. 

 We  configure the model  for QLoRA fine-tuning with the following parameters:
\begin{itemize}
\item Low Rank: Set to 64 to balance between performance and computational efficiency.
\item Alpha: Set to 16 to scale the low-rank updates.
\item Dropout: Set to 0.05 to prevent overfitting during training.
\item Adapter Layers: We add adapter layers to the Query, Key, and Value and Feedforward Weights of the Transformer layers of the Phi-2 model.
% Bias: Set to "none" as bias terms are not updated.
% Task Type: Defined as "CAUSAL_LM" indicating the causal language modeling task.
% Target Modules: The specific modules within the model targeted for low-rank adaptation include "q_proj", "    k_proj", "v_proj", and "dense".
\end{itemize}

\subsubsection{LoRa}
The use of QLoRA finetuning reduces the memory requirement during training. However, the training time is considerably increased due to the quantization and dequantization operations being performed during QLoRA finetuning. In order to investigate the influence of the finetuning objective on model's performance on MCQs, we used LoRA since it offers faster training time compared to QLoRA. We compared the model's performance when the training objective only considers the answers versus when the entire prompt and answer are considered in the objective.

The original cross-entropy loss for next token prediction is defined as:

\[
\mathcal{L} = - \sum_{t=1}^{T} y_t \log(\hat{y}_t)
\]

where:
\begin{itemize}
    \item \( T \) is the total number of tokens in the sequence (including both the question/prompt and the answer).
    \item \( y_t \) is the actual token at position \( t \) (one-hot encoded).
    \item \( \hat{y}_t \) is the predicted probability distribution over the vocabulary for token \( t \).
\end{itemize}

To focus only on answer generation, we introduce a masking vector \( m_t \) such that:
\[
m_t = 
\begin{cases} 
0 & \text{if } t \in \{1, 2, \ldots, Q\} \\
1 & \text{if } t \in \{Q+1, Q+2, \ldots, T\}
\end{cases}
\]

The modified cross-entropy loss, which we call question-masked loss \( \mathcal{L}_{\text{masked}} \) is then:

\[
\mathcal{L}_{\text{masked}} = - \sum_{t=1}^{T} m_t \cdot y_t \log(\hat{y}_t)
\]

This loss function ensures that only the tokens corresponding to the answer part of the sequence contribute to the overall loss, effectively masking out the contributions from the question/prompt part.

\subsection{Option Batch-Shuffle Trick}\label{batchshuf}

Recent research~\cite{maatouk2024teleqna, pezeshkpour2023large, zheng2023large} has unveiled a significant bias in LLMs  when answering multiple-choice questions (MCQs). These models exhibit a strong sensitivity to the order of options, often selecting specific answer positions regardless of the content. This phenomenon, termed "selection bias," stems from the models' tendency to assign higher probabilities to certain option labels (like "A" or "B" in options ["A", "B", "C", "D" and "E"] ). Consequently, LLMs may prioritize these options even when logically incorrect, undermining the reliability of their performance on MCQ assessments. To avert this we employ a trick where we create multiple prompts for a question,
with each prompt having a different option order. The correct answer is thus determined by chosing the most selected answer by the model after observing the answers generated for all created prompts. Given that we have to permute these options to obtain all possible option ordering, the complexity of doing so is $O(n!)$ where $n$ is the number of options present in an MCQ. This complexity significantly increases the inference time for a single question. For example, when an MCQ has 4 options, we create $4! = 24$ prompts instead of 1 prompt. For 5 options, we create $5! = 120$ prompts. To reduce this complexity, we randomly sample $k$ prompts from the n! prompts, create a batch of $k$ prompts and generate answers for the batch using the Phi-2 model. Using \(k\) prompts instead of \(n!\) reduces the complexity from \(O(n!)\) to \(O(k)\), where \(k \ll n!\). The model generates answers for these \(k\) prompts in a single batch, thus significantly reducing inference time while still benefiting from diverse option orderings. The selection of the most frequent answer from these \(k\) prompts can be described as:

\begin{equation}
\hat{a} = \arg\max_{a \in A} \sum_{i=1}^{k} \mathbb{I}(a_i = a)
\end{equation}

where \(\mathbb{I}(\cdot)\) is the indicator function, \(a_i\) is the answer chosen by the model for the \(i\)-th prompt, and \(A\) is the set of all possible answers.

 We call this the '\textit{batch-shuffle trick}'. Using the batch-shuffle trick, we noticed over 6\%  at inference time and this increased to about 10\% when we include the trick into the training phase. In the training phase, we only shuffle the options at the end of each epoch and do not use any explicit sampling. We hypothesize further improvement in performance as we increase $k$, the number of samples from n! prompts of an MCQ. In our case, we find $k=20$ to be a good balance between efficiency and accuracy. 
 
 % \tdCarlee{Minor point, but some instances of k are not in math mode in this section. Also, if there is space, would be good to show how the accuracy changes as you vary $k$, since this is a new hyperparameter that you introduce}

% Describe the idea of permuting the options of a single question to generate several prompts with different option orderings and how the correct option is chosen using the model.

\section{Evaluation}
We evaluated our approach using a subset of the TeleQnA dataset containing only two categories; Standards Specifications and Standards Overview. Matoouk \emph{et al}~\cite{maatouk2024teleqna}
showed that GPT-3.5 and GPT-4 performed better in other question categories than in these two categories. The dataset, obtained from the "Specializing Large Language Models for Telecom Networks by ITU AI/ML in 5G Challenge" on Zindi \cite{zindi2024telecom} contains a train set of 1461 questions, a public test set 366 questions, and a private test set of 2000 questions. Evaluation results, as obtained from the leaderboard, are reported for the private test set. The train set was used for fine-tuning purposes. Additionally, we used the technical documents provided by the challenge as external knowledge sources for RAG. 

We compare the performance of the base Phi-2 model and its performance with RAG, with fine-tuning and with the batch-shuffle trick both at inference and at training time. For the Falcon7B, we compare the base model performance and its performance with RAG and with the options excluded in the prompt. 

% \tdCarlee{Mention the ablation baselines you compare against here: Phi-2, Phi-2 + RAG, etc.}

\subsection{Phi-2 model}
Table~\ref{tab:base-phi-2-performance} shows the performance of the Phi-2 model on the private test sets as obtained from the submissions on Zindi. The accuracy score measures the percentage of correctly answered questions. The base Phi-2 model has an accuracy of 42.07\%, and this accuracy was increased to 66.39\% with the introduction of RAG. The result obtained with RAG was further improved by fine-tuning the model using the training set. With fine-tuning, the accuracy increased to 76.90\%. Using the batch-shuffle trick (see section \ref{batchshuf}) with the fine-tuned Phi-2 model, we obtained an accuracy of 81.65\% which is a 6.18\% increase over the fine-tuned model and 84.65\% when we shuffle the options during training. While fine-tuning the Phi-2 model on the training data, we discovered that the model's performance stops improving after certain epochs of training. To investigate this we modified the objective (next token prediction) to only account for answer generation by masking the part of the cross entropy loss associated with question/prompt prediction. 

\begin{figure}
    \centering
    \includegraphics[width=1.0\linewidth]{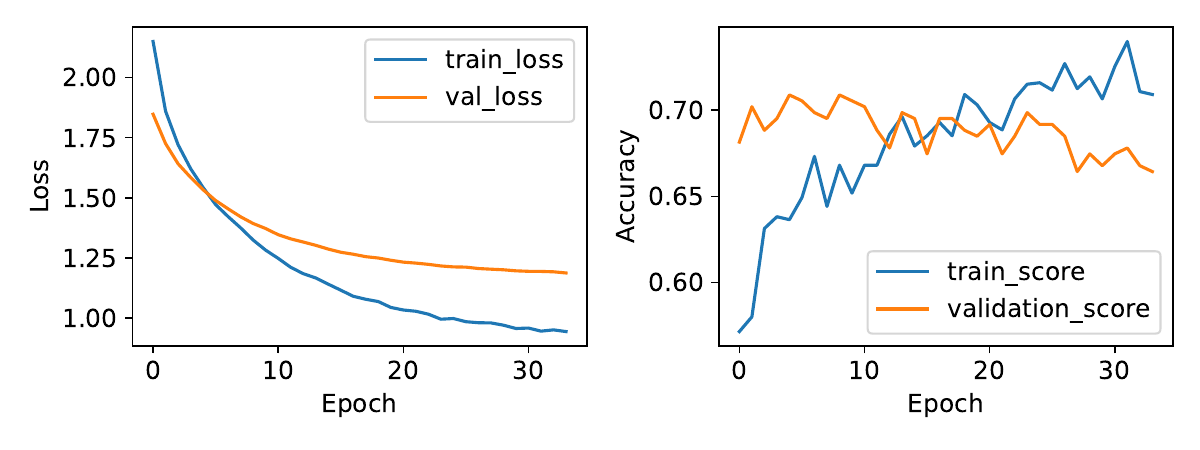}
    \caption{Loss and accuracy scores when fine-tuning the Phi-2 model with the standard next-token prediction objective}
    \label{fig:original-objective}
\end{figure}

\begin{figure}
    \centering
    \includegraphics[width=1.0\linewidth]{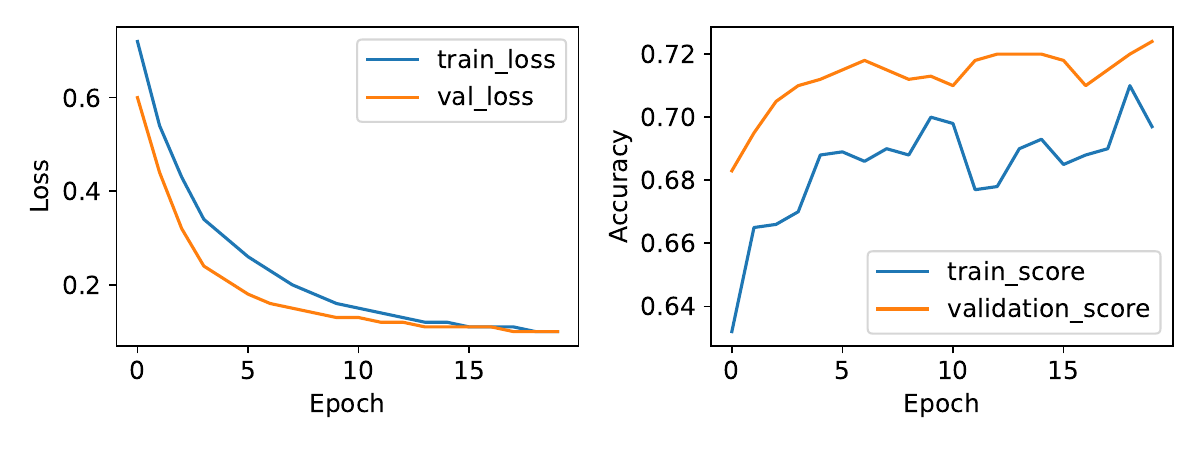}
    \caption{Loss and accuracy scores when fine-tuning the Phi-2 model and considering only the answers in the next-token prediction objective}
    \label{fig:question-loss-masking}
\end{figure}

Using a train-validation split (20\% validation) while fine-tuning, it is expected that the validation accuracy increases as the validation loss decreases. However, Figure~\ref{fig:original-objective} shows no improvement in the validation accuracy even as the validation loss decreases. We suspected that this decrease in validation loss results from the model getting better at predicting the question and not the answers. Figure~\ref{fig:question-loss-masking} shows the result obtained when the loss associated with questions is masked out, allowing the training objective to focus only on the answers. Figure~\ref{fig:question-loss-masking} shows that the validation accuracy increases as the validation loss decreases. We, therefore, hypothesize that focusing solely on the answers during fine-tuning may yield better results. Validating this hypothesis is an interesting area of future work to be explored with additional experiments.
%\tdCarlee{why is this insufficient? You might phrase this differently, e.g., ``We conjecture that focusing solely on the answers during fine-tuning may yield better results. Validating this conjecture is an interesting area of future work.''}

\begin{table}[h!]
\caption{Performance of Phi-2 model on the test dataset as obtained from submissions to Zindi}
\centering
\begin{tabular}{|p{0.5\linewidth}|p{0.3\linewidth}|}
    \hline
    \textbf{Method} & \textbf{Test Accuracy (\%)} \\
    \hline
    GPT-3.5 &  56.97**  \\
    \hline
    GPT-4 &  64.78** \\
    \hline
    Phi-2 &  42.07 \\
    % \hline
    % Phi-2 + RAG & 56.63** \\
    \hline
    Phi-2 + RAG & 66.39 (Ours)\\
    \hline
    Phi-2 + RAG + Fine-tuning & 76.90 \\
    \hline
    Phi-2 + RAG + Fine-tuning + Inference batch-shuffle & 81.65\\
    \hline
    Phi-2 + RAG + Fine-tuning + Inference \& Train batch-shuffle (\textbf{QMOS}) & \textbf{84.65} \\ 
    \hline
\end{tabular}
\label{tab:base-phi-2-performance}
\end{table}
\textcolor{black}{** Results from \cite{piovesan2024telecomlanguagemodelslarge} which used a different test set.
}

\subsection{Falcon-7B}

The performance of the Falcon-7B model is summarized in Table \ref{tab:falcon-performance}
\begin{table}[h!]
\caption{Performance of Falcon7B model on the test dataset as obtained from submissions to Zindi}
    \centering
    \begin{tabular}{|c|c|}
        \hline
        \textbf{Method} & \textbf{Accuracy (\%)}  \\
        \hline
        Baseline Falcon7B &  24.51 \\
        \hline
        Falcon7B + RAG & 36.61 \\
        \hline
        Falcon7B + RAG + No Options & \textbf{49.30} \\
        \hline
        % Baseline Falcon7B & 0.00 \\
        % \hline 
    \end{tabular}
    
    \label{tab:falcon-performance}
\end{table}

For the Baseline 7B model, when prompted with the options we notice that in some cases the model does not output the options but just some unrelated texts.In that case, we randomly 
choose an option. This yielded an accuracy of 24.51\%. Adding the contexts from the RAG further enhanced the score to 36.61\%. Finally, by removing the options and allowing the model to generate the answer freely. We then used the embedding of the generated answer with the embedding of the options to select the right option using a cosine similarity metric.  With this strategy the model was able to achieve an accuracy of \textbf{49.93\%}, which is significantly higher than the baseline of  \textbf{24.51\%}.
We note that the baseline Falcon-7B model is not fine-tuned considering it has more parameters (7B) than Phi-2 (2.7B). Since fine-tuning will increase the computational cost, we strictly rely on the effectiveness of prompting and RAG systems in a bid to improve its baseline performance on question answering in telecommunications domain. 

% we noticed that the mode

\section{Conclusion}

In this research, we have presented a comprehensive approach to addressing multiple-choice questions (MCQs) in the telecommunications domain using small, open-source language models within an enhanced Retrieval-Augmented Generation (RAG) framework. Our study demonstrates that small models such as Phi-2 and Falcon-7B, when combined with advanced techniques like LoRA fine-tuning, diversified embedding models for RAG, innovative prompt engineering, and batch-shuffle trick can achieve competitive performance compared to larger, proprietary models like GPT-3.5, while significantly reducing computational costs.Future work will involve fine-tuning the embedding models (used for RAG) for the telecommunication domain and also further investigate the performance of the proposed QMOS framework on the other language models and MCQs datasets.

\bibliographystyle{IEEEtran}
\bibliography{references}

\end{document}